# Evolved Policy Gradients

Rein Houthooft [1]   Richard Y. Chen [1]   Phillip Isola [1 2 3]   Bradly C. Stadie [2]   Filip Wolski [1]
Jonathan Ho [1 2]   Pieter Abbeel [1 2]

## Abstract

We propose a metalearning approach for learning gradient-based reinforcement learning (RL) algorithms. The idea is to evolve a differentiable loss function, such that an agent, which optimizes its policy to minimize this loss, will achieve high rewards. The loss is parametrized via temporal convolutions over the agent's experience. Because this loss is highly flexible in its ability to take into account the agent's history, it enables fast task learning. Empirical results show that our evolved policy gradient algorithm (EPG) achieves faster learning on several randomized environments compared to an off-the-shelf policy gradient method. We also demonstrate that EPG's learned loss can generalize to out-of-distribution test time tasks, and exhibits qualitatively different behavior from other popular metalearning algorithms.

## 1. Introduction

When a human learns to solve a new control task, such as playing the violin, they immediately have a feel for what to try. At first, they may try a quick, rough stroke, and, producing a screech, will intuitively know this was the wrong thing to do. Just by listening to the sounds they produce, they will have a sense of whether or not they are making progress toward the goal. Effectively, humans have access to very well shaped internal reward functions, derived from prior experience on other motor tasks, or perhaps from listening to and playing other musical instruments (36; 49).

In contrast, most current reinforcement learning (RL) agents approach each new task de novo. Initially, they have no notion of what actions to try out, nor which outcomes are desirable. Instead, they rely entirely on external reward signals to guide their initial behavior. Coming from such a blank slate, it is no surprise that RL agents take far longer than humans to learn simple skills (21).

[1]OpenAI [2]UC Berkeley [3]MIT. Correspondence to: Rein Houthooft <rein.houthooft@openai.com>.

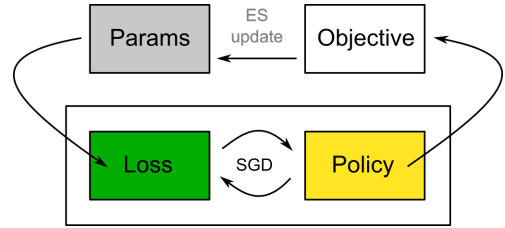

Figure 1: High-level overview of our approach. The method consists of an inner and outer optimization loop. The inner loop (boxed) optimizes the agent's policy against a loss provided by the outer loop, using gradient descent. The outer loop optimizes the parameters of the loss function, such that the optimized inner-loop policy achieves high performance on an arbitrary task, such as solving a control task of interest. The evolved loss $L$ can be viewed as a surrogate whose gradient is used to update the policy, which is similar in spirit to policy gradients, lending the name "evolved policy gradients".

Our aim in this paper is to devise agents that have a prior notion of what constitutes making progress on a novel task. Rather than encoding this knowledge explicitly through a learned behavioral policy, we encode it implicitly through a learned loss function. The end goal is agents that can use this loss function to learn quickly on a novel task.

This approach can be seen as a form of metalearning, in which we learn a learning algorithm. Rather than mining rules that generalize across data points, as in traditional machine learning, metalearning concerns itself with devising algorithms that generalize across tasks, by infusing prior knowledge of the task distribution (12).

Our method consists of two optimization loops. In the inner loop, an agent learns to solve a task, sampled from a particular distribution over a family of tasks. The agent learns to solve this task by minimizing a loss function provided by the outer loop. In the outer loop, the parameters of the loss function are adjusted so as to maximize the final returns achieved after inner loop learning. Figure 1 provides a high-level overview of this approach.

Although the inner loop can be optimized with stochastic gradient descent (SGD), optimizing the outer loop presents substantial difficulty. Each evaluation of the outer objective requires training a complete inner-loop agent, and this objective cannot be written as an explicit function of the loss parameters we are optimizing over. Due to the lack of easily exploitable structure in this optimization problem, we turn to evolution strategies (ES) (35; 46; 16; 37) as a blackbox optimizer. The evolved loss $L$ can be viewed as a surrogate loss (43; 44) whose gradient is used to update the policy, which is similar in spirit to policy gradients, lending the name "evolved policy gradients".

In addition to encoding prior knowledge, the learned loss offers several advantages compared to current RL methods. Since RL methods optimize for short-term returns instead of accounting for the complete learning process, they may get stuck in local minima and fail to explore the full search space. Prior works add auxiliary reward terms that emphasize exploration (8; 19; 32; 56; 6; 33) and entropy loss terms (31; 42; 15; 26). These terms are often traded off using a separate hyperparameter that is not only task-dependent, but also dependent on which part of the state space the agent is visiting. As such, it is unclear how to include these terms into the RL algorithm in a principled way.

Using ES to evolve the loss function allows us to optimize the true objective, namely the final trained policy performance, rather than short-term returns. Our method also improves on standard RL algorithms by allowing the loss function to be adaptive to the environment and agent history, leading to faster learning and the potential for learning without external rewards. EPG can in theory be combined with policy initialization metalearning algorithms, such as MAML (11), since EPG imposes no restriction on the policy it optimizes.

There has been a flurry of recent work on metalearning policies, e.g., (10; 59; 11; 25), and it is worth asking why metalearn the loss as opposed to directly metalearning the policy? Our motivation is that we expect loss functions to be the kind of object that may generalize very well across substantially different tasks. This is certainly true of hand-engineered loss functions: a well-designed RL loss function, such as that in (45), can be very generically applicable, finding use in problems ranging from playing Atari games to controlling robots (45). In Section 4, we find evidence that a loss learned by EPG can train an agent to solve a task *outside the distribution* of tasks on which EPG was trained. This generalization behavior differs qualitatively from MAML (11) and RL$^2$ (10), methods that directly metalearn policies, providing initial indication of the generalization potential of loss learning.

Our contributions include the following:

- Formulating a metalearning approach that learns a differentiable loss function for RL agents, called EPG.

- Optimizing the parameters of this loss function via ES, overcoming the challenge that final returns are not explicit functions of the loss parameters.

- Designing a loss architecture that takes into account agent history via temporal convolutions.

- Demonstrating that EPG produces a learned loss that can train agents faster than an off-the-shelf policy gradient method.

- Showing that EPG's learned loss can generalize to *out-of-distribution* test time tasks, exhibiting qualitatively different behavior from other popular metalearning algorithms.

We set forth the notation in Section 2. Section 3 explains the main algorithm and Section 4 shows its results on several randomized continuous control environments. In Section 5, we compare our methods with the most related ideas in literature. We conclude this paper with a discussion in Section 6. An implementation of EPG is available at

http://github.com/openai/EPG.

## 2. Notation and Background

We model reinforcement learning (54) as a Markov decision process (MDP), defined as the tuple $\mathcal{M} = (\mathcal{S}, \mathcal{A}, T, R, p_0, \gamma)$, where $\mathcal{S}$ and $\mathcal{A}$ are the state and action space. The transition dynamic $T : \mathcal{S} \times \mathcal{A} \times \mathcal{S} \mapsto \mathbb{R}_+$ determines the distribution of the next state $s_{t+1}$ given the current state $s_t$ and the action $a_t$. $R : \mathcal{S} \times \mathcal{A} \mapsto \mathbb{R}$ is the reward function and $\gamma \in (0, 1)$ is a discount factor. $p_0$ is the distribution of the initial state $s_0$. An agent's policy $\pi : \mathcal{S} \mapsto \mathcal{A}$ generates an action after observing a state.

An episode $\tau \sim \mathcal{M}$ with horizon $H$ is a sequence $(s_0, a_0, r_0, \ldots, s_H, a_H, r_H)$ of state, action, and reward at each timestep $t$. The discounted episodic return of $\tau$ is defined as $R_\tau = \sum_{t=0}^{H} \gamma^t r_t$, which depends on the initial state distribution $p_0$, the agent's policy $\pi$, and the transition distribution $T$. The expected episodic return given agent's policy $\pi$ is $\mathbb{E}_\pi[R_\tau]$. The optimal policy $\pi^*$ maximizes the expected episodic return

$$\pi^* = \arg\max_\pi \mathbb{E}_{\tau \sim \mathcal{M}, \pi}[R_\tau].$$

In high-dimensional reinforcement learning settings, the policy $\pi$ is often parametrized using a deep neural network $\pi_{\boldsymbol{\theta}}$ with parameters $\boldsymbol{\theta}$. The goal is to solve for $\boldsymbol{\theta}^*$ that attains the highest expected episodic return

$$\boldsymbol{\theta}^* = \arg\max_{\boldsymbol{\theta}} \mathbb{E}_{\tau \sim \mathcal{M}, \pi_{\boldsymbol{\theta}}}[R_\tau]. \tag{1}$$

This objective can be optimized via policy gradient methods (60; 55) by stepping in the direction of $\mathbb{E}[R_\tau \nabla \log \pi(\tau)]$. This gradient can be transformed into a surrogate loss function (43; 44)

$$L_{\text{pg}} = \mathbb{E}[R_\tau \log \pi(\tau)] = \mathbb{E}\left[R_\tau \sum_{t=0}^{H} \log \pi(a_t|s_t)\right], \quad (2)$$

such that the gradient of $L_{\text{pg}}$ equals the policy gradient. Through variance reduction techniques including actor-critic algorithms (20), the loss function $L_{\text{pg}}$ is often changed into

$$L_{\text{ac}} = \mathbb{E}\left[\sum_{t=0}^{H} A(s_t, a_t) \log \pi(a_t|s_t)\right], \quad (3)$$

that is, the log-probability of taking action $a_t$ at state $s_t$ is multiplied by an advantage function $A(s_t, a_t)$ (4).

However, this procedure remains limited since it relies on a particular form of discounting the returns, and taking a fixed gradient step with respect to the policy. Our approach learns a loss rather than using a hand-defined function such as $L_{ac}$. Thus, it may be able to discover more effective surrogates for making fast progress toward the ultimate objective of maximizing final returns.

## 3. Methodology

Our metalearning approach aims to learn a loss function $L_\phi$ that outperforms the usual policy gradient surrogate loss (43). This loss function consists of temporal convolutions over the agent's recent history. In addition to internalizing environment rewards, this loss could, in principle, have several other positive effects. For example, by examining the agent's history, the loss could incentivize desirable extended behaviors, such as exploration. Further, the loss could perform a form of system identification, inferring environment parameters and adapting how it guides the agent as a function of these parameters (e.g., by adjusting the effective learning rate of the agent).

The loss function parameters $\phi$ are evolved through ES and the loss trains an agent's policy $\pi_\theta$ in an on-policy fashion via stochastic gradient descent.

### 3.1. Metalearning Objective

In our metalearning setup, we assume access to a distribution $p(\mathcal{M})$ over MDPs. Given a sampled MDP $\mathcal{M}$, the inner loop optimization problem is to minimize the loss $L_\phi$ with respect to the agent's policy $\pi_\theta$:

$$\theta^* = \arg\min_\theta \mathbb{E}_{\tau \sim \mathcal{M}, \pi_\theta}[L_\phi(\pi_\theta, \tau)]. \quad (4)$$

Note that this is similar to the usual RL objectives (Eqs. (1) (2) (3)), except that we are optimizing a learned loss

---

**Algorithm 1:** Evolved Policy Gradients (EPG)

1 **[Outer Loop] for** epoch $e = 1, \ldots, E$ **do**
2     Sample $\epsilon_v \sim \mathcal{N}(\mathbf{0}, \mathbf{I})$ and calculate the loss parameter $\phi + \sigma\epsilon_v$ for $v = 1, \ldots, V$
3     Each worker $w = 1, \ldots, W$ gets assigned noise vector $\lceil wV/W \rceil$ as $\epsilon_w$
4     **for** each worker $w = 1, \ldots, W$ **do**
5        Sample MDP $\mathcal{M}_w \sim p(\mathcal{M})$
6        Initialize buffer with $N$ zero tuples
7        Initialize policy parameter $\theta$ randomly
8        **[Inner Loop] for** step $t = 1, \ldots, U$ **do**
9           Sample initial state $s_t \sim p_0$ if $\mathcal{M}_w$ needs to be reset
10          Sample action $a_t \sim \pi_\theta(\cdot|s_t)$
11          Take action $a_t$ in $\mathcal{M}_w$ and receive $r_t, s_{t+1}$, and termination flag $d_t$
12          Add tuple $(s_t, a_t, r_t, d_t)$ to buffer
13          **if** $t \mod M = 0$ **then**
14             With loss parameter $\phi + \sigma\epsilon_w$, calculate losses $L_i$ for steps $i = t - M, \ldots, t$ using buffer tuples $i - N, \ldots, i$
15             Sample minibatches mb from last $M$ steps shuffled, compute $L_{\text{mb}} = \sum_{j \in \text{mb}} L_j$, and update the policy parameter $\theta$ and memory parameter (Eq. (6))
16     In $\mathcal{M}_w$, using the trained policy $\pi_\theta$, sample several trajectories and compute the mean return $R_w$
17     Update the loss parameter $\phi$ (Eq. (7))
18 **Output:** Loss $L_\phi$ that trains $\pi$ from scratch according to the inner loop scheme, on MDPs from $p(\mathcal{M})$

---

$L_\phi$ rather than directly optimizing the expected episodic return $\mathbb{E}_{\mathcal{M}, \pi_\theta}[R_\tau]$ or other surrogate losses. The outer loop objective is to learn $L_\phi$ such that an agent's policy $\pi_{\theta^*}$ trained with the loss function achieves high expected returns in the MDP distribution:

$$\phi^* = \arg\max_\phi \mathbb{E}_{\mathcal{M} \sim p(\mathcal{M})} \mathbb{E}_{\tau \sim \mathcal{M}, \pi_{\theta^*}}[R_\tau]. \quad (5)$$

### 3.2. Algorithm

The final episodic return $R_\tau$ of a trained policy $\pi_{\theta^*}$ cannot be represented as an explicit function of the loss function $L_\phi$. Thus we cannot use gradient-based methods to directly solve Eq. (5). Our approach, summarized in Algorithm 1, relies on evolution strategies (ES) to optimize the loss function in the outer loop.

As described by Salimans et al. (37), ES computes the gra-

**Algorithm 2:** EPG test-time training

1  **[Input]**: learned loss function $L_\phi$ from EPG, MDP $\mathcal{M}$
2  Initialize buffer with $N$ zero tuples
3  Initialize policy parameter $\boldsymbol{\theta}$ randomly
4  **for** step $t = 1, \ldots, U$ **do**
5      Sample initial state $s_t \sim p_0$ if $\mathcal{M}$ needs to be reset
6      Sample action $a_t \sim \pi_{\boldsymbol{\theta}}(\cdot|s_t)$
7      Take action $a_t$ in $\mathcal{M}$, receive $r_t, s_{t+1}$, and termination flag $d_t$
8      Add tuple $(s_t, a_t, r_t, d_t)$ to buffer
9      **if** $t \bmod M = 0$ **then**
10         Calculate losses $L_i$ for steps $i = t - M, \ldots, t$ using buffer tuples $i - N, \ldots, i$
11         Sample minibatches mb from last $M$ steps shuffled, compute $L_{\mathrm{mb}} = \sum_{j \in \mathrm{mb}} L_j$, and update the policy parameter $\boldsymbol{\theta}$ and memory parameter (Eq. (6))
12 **[Output]**: A trained policy $\pi_{\boldsymbol{\theta}}$ for MDP $\mathcal{M}$

dient of a function $F(\phi)$ according to

$$\nabla_{\boldsymbol{\phi}} \mathbb{E}_{\boldsymbol{\epsilon} \sim \mathcal{N}(0,I)} F(\boldsymbol{\phi} + \sigma \boldsymbol{\epsilon}) = \frac{1}{\sigma} \mathbb{E}_{\boldsymbol{\epsilon} \sim \mathcal{N}(0,I)} F(\boldsymbol{\phi} + \sigma \boldsymbol{\epsilon}) \boldsymbol{\epsilon}.$$

Similar formulations also appear in prior works including (52; 47; 27). In our case, $F(\boldsymbol{\phi}) = \mathbb{E}_{\mathcal{M} \sim p(\mathcal{M})} \mathbb{E}_{\tau \sim \mathcal{M}, \pi_{\boldsymbol{\theta}^*}}[R_\tau]$ (Eq. (5)). Note that the dependence on $\phi$ comes through $\boldsymbol{\theta}^*$ (Eq. (4)).

Step by step, the algorithm works as follows. At the start of each epoch in the outer loop, for $W$ inner-loop workers, we generate $V$ standard multivariate normal vectors $\boldsymbol{\epsilon}_v \in \mathcal{N}(0, \mathrm{I})$ with the same dimension as the loss function parameter $\phi$, assigned to $V$ sets of $W/V$ workers. As such, for the $w$-th worker, the outer loop assigns the $\lceil wV/W \rceil$-th perturbed loss function

$$L_w = L_{\boldsymbol{\phi} + \sigma \boldsymbol{\epsilon}_v} \text{where } v = \lceil wV/W \rceil$$

with perturbed parameters $\boldsymbol{\phi} + \sigma \boldsymbol{\epsilon}_v$ and $\sigma$ as the standard deviation.

Given a loss function $L_w, w \in \{1, \ldots, W\}$, from the outer loop, each inner-loop worker $w$ samples a random MDP from the task distribution, $\mathcal{M}_w \sim p(\mathcal{M})$. The worker then trains a policy $\pi_{\boldsymbol{\theta}}$ in $\mathcal{M}_w$ over $U$ steps of experience. Whenever a termination signal is reached, the environment resets with state $s_0$ sampled from the initial state distribution $p_0(\mathcal{M}_w)$. Every $M$ steps the policy is updated through SGD on the loss function $L_w$, using minibatches sampled from the steps $t - M, \ldots, t$:

$$\boldsymbol{\theta} \leftarrow \boldsymbol{\theta} - \delta_{\mathrm{in}} \cdot \nabla_{\boldsymbol{\theta}} L_w(\pi_{\boldsymbol{\theta}}, \tau_{t-M,\ldots,t}). \quad (6)$$

At the end of the inner-loop training, each worker returns the final return $R_w$[1] to the outer loop. The outer-loop aggregates the final returns $\{R_w\}_{w=1}^W$ from all workers and updates the loss function parameter $\phi$ as follows:

$$\boldsymbol{\phi} \leftarrow \boldsymbol{\phi} + \delta_{\mathrm{out}} \cdot \frac{1}{V\sigma} \sum_{v=1}^V F(\boldsymbol{\phi} + \sigma \boldsymbol{\epsilon}_v) \boldsymbol{\epsilon}_v, \quad (7)$$

where

$$F(\boldsymbol{\phi} + \sigma \boldsymbol{\epsilon}_v) = \frac{R_{(v-1)*W/V+1} + \cdots + R_{v*W/V}}{W/V}.$$

As a result, each perturbed loss function $L_v$ is evaluated on $W/V$ randomly sampled MDPs from the task distribution using the final returns. This achieves variance reduction by preventing the outer-loop ES update from promoting loss functions that are assigned to MDPs that consistently generate higher returns. Note that the actual implementation calculates each loss function's relative rank for the ES update. Algorithm 1 outputs a learned loss function $L_\phi$ after $E$ epochs of ES updates.

At test time, we evaluate the learned loss function $L_\phi$ produced by Algorithm 1 on a test MDP $\mathcal{M}$ by training a policy from scratch. The test-time training schedule is the same as the inner loop of Algorithm 1 and we summarize it in Algorithm 2.

### 3.3. Architecture

The agent is parametrized using an MLP policy with observation space $\mathcal{S}$ and action space $\mathcal{A}$. The loss has a memory unit to assist learning in the inner loop. This memory unit is a single-layer neural network to which an invariable input vector of ones is fed. As such, it is essentially a layer of bias terms. Since this network has a constant input vector, we can view its weights as a very simple form of memory to which the loss can write via emitting the right gradient signals. An experience buffer stores the agent's $N$ most recent experience steps, in the form of a list of tuples $(s_t, a_t, r_t, d_t)$, with $d_t$ the trajectory termination flag. Since this buffer is limited in the number of steps it stores, the memory unit might allow the loss function to store information over a longer period of time.

The loss function $L_\phi$ consists of temporal convolutional layers which generate a context vector $f_{\mathrm{context}}$, and dense layers, which output the loss. The architecture is depicted in Figure 2.

At step $t$, the dense layers output the loss $L_t$ by taking a batch of $M$ sequential samples

$$\{s_i, a_i, d_i, \mathrm{mem}, f_{\mathrm{context}}, \pi_{\boldsymbol{\theta}}(\cdot|s_i)\}_{i=t-M}^t, \quad (8)$$

where $M < N$ and we augment each transition with the memory output mem, a context vector $f_{\mathrm{context}}$ generated from the loss's temporal convolutional layers, and the policy distribution $\pi_{\boldsymbol{\theta}}(\cdot|s_i)$. In continuous action space, $\pi_{\boldsymbol{\theta}}$ is a Gaussian policy, i.e., $\pi_{\boldsymbol{\theta}}(\cdot|s_i) = \mathcal{N}(\cdot; \mu(s_i; \boldsymbol{\theta}_0), \Sigma)$, with $\mu(s_i; \boldsymbol{\theta}_0)$ the MLP output and $\Sigma$ a learnable parameter vector. The policy parameter vector is defined as $\boldsymbol{\theta} = [\boldsymbol{\theta}_0, \Sigma]$.

---

[1]More specifically, the average return over 3 sampled trajectories using the final policy for worker $w$.

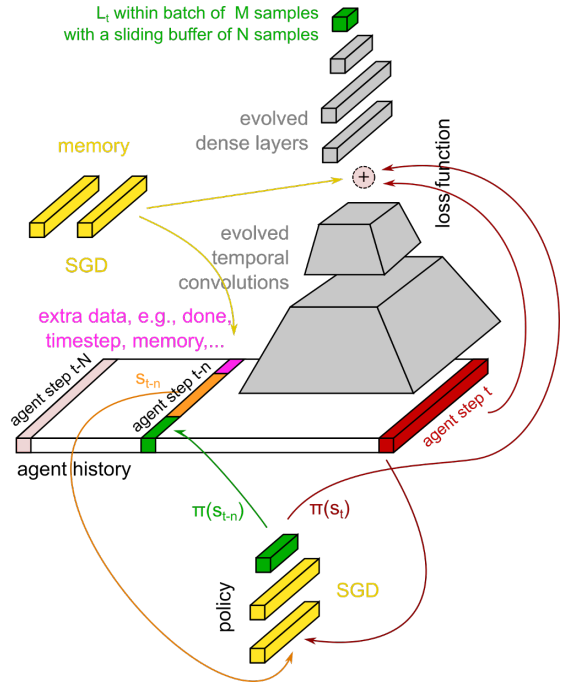

Figure 2: Architecture of a loss computed for timestep $t$ within a batch of $M$ sequential samples (from $t-M$ to $t$), using temporal convolutions over a buffer of size $N$ (from $t-N$ to $t$), with $M \leq N$: dense net on the bottom is the policy $\pi(s)$, taking as input the observations (orange), while outputting action probabilities (green). The green block on the top represents the loss output. Gray blocks are evolved, yellow blocks are updated through SGD.

To generate the context vector, we first augment each transition in the buffer with the output of the memory unit mem and the policy distribution $\pi_{\boldsymbol{\theta}}(\cdot|s_i)$ to obtain a set

$$\{s_i, a_i, d_i, \text{mem}, \pi_{\boldsymbol{\theta}}(\cdot|s_i)\}_{i=t-N}^{t}. \qquad (9)$$

We stack these items sequentially into a matrix and the temporal convolutional layers take it as input and output the context vector $f_{\text{context}}$. The memory unit's parameters are updated via gradient descent at each inner-loop update (Eq. (6)).

Note that both the temporal convolution layers and the dense layers do not observe the environment rewards directly. However, in cases where the reward cannot be fully inferred from the environment, such as the DirectionalHopper environment we will examine in Section 4.2, we add rewards $r_i$ to the set of inputs in Eqs. (8) and (9). In fact, any information that can be obtained from the environment could be added as an input to the loss function, e.g., exploration signals, the current timestep number, etc, and we leave further such extensions as future work.

In practice, to bootstrap the learning process, we add to $L_{\phi}$ a guidance policy gradient surrogate loss signal $L_{\text{pg}}$, such as the REINFORCE (60) or PPO (45) surrogate loss function, making the total loss

$$\hat{L}_{\phi} = (1-\alpha)L_{\phi} + \alpha L_{\text{pg}}, \qquad (10)$$

and anneal $\alpha$ from 1 to 0 over a finite number of outer-loop epochs. As such, learning is first derived mostly from the well-structured $L_{\text{pg}}$, while over time $L_{\phi}$ takes over and drives learning completely after $\alpha$ has been annealed to 0.

## 4. Experiments

We apply our method to several randomized continuous control MuJoCo environments (5; 34; 9), namely RandomHopper and RandomWalker (with randomized gravity, friction, body mass, and link thickness), RandomReacher (with randomized link lengths), DirectionalHopper and DirectionalHalfCheetah (with randomized forward/backward reward function), GoalAnt (reward function based on the randomized target location), and Fetch (randomized target location). We describe these environments in detail in Appendix A. These environments are chosen because they require the agent to identify a randomly sampled environment at test time via exploratory behavior. Examples of the randomized Hopper environments are shown in Figure 4 and the Fetch environment in Figure 3. The plots in this section show the mean value of 20 test-time training curves as a solid line, while the shaded area represents the interquartile range. The dotted lines plot 5 randomly sampled curves.

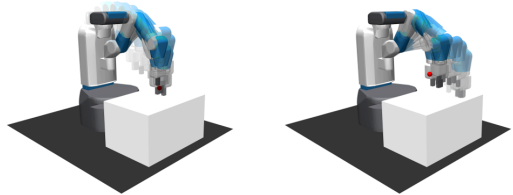

Figure 3: Examples of learning to reach random targets in the Fetch environment

**Implementation details** In our experiments, the temporal convolutional layers of the loss function has 3 layers. The first layer has a kernel size of 8, stride of 7, and outputs 10 channels. The second layer has a kernel of 4, stride of 2, and outputs 10 channels. The third layer is fully-connected with 32 output units. Leaky ReLU activation is applied to each convolutional layer. The fully-connected component takes as input the trajectory features from the convolutional component concatenated with state, action, termination signal, and policy output, as well as reward in experiments in which reward is observed. It has 1 hidden layer with 16 hidden units and leaky ReLU activation, followed by an output layer. The buffer size is $N \in \{512, 1024\}$. The agent's MLP policy has 2 hidden layers of 64 units with tanh activation. The memory unit is a 32-unit single layer with tanh activation.

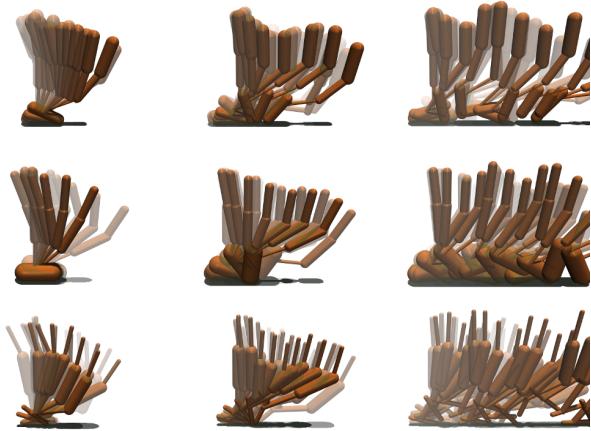

Figure 4: Example of learning to hop forward from a randomly initialized policy in RandomHopper environments with randomized morphology and physics parameters. Each row is a different environment randomization, while from left to right, trajectories are recorded as learning progresses.

We use $W = 256$ inner-loop workers in Algorithm 1, combined with $V = 64$ ES noise vectors. The loss function is evolved over 5000 epochs, with $\alpha$, as in Eq. (10), annealed linearly from 1 to 0 over the first 500 epochs. The off-the-shelf PG algorithm (PPO) was moderately tuned to perform well on these tasks, however, it is important to keep in mind that these methods inherently have trouble optimizing when the number of samples drawn for each policy update batch is low. EPG's inner loop update frequency is set to $M \in \{32, 64, 128\}$ and the inner loop length is $U \in \{64 \times M, 128 \times M, 256 \times M, 512 \times M\}$. At every EPG inner loop update, the policy and memory parameters are updated by the learned loss function using shuffled minibatches of size 32 within each set of $M$ most recent transition steps in the replay buffer, going over each step exactly once. We tabulate the hyperparameters for each randomized environment in Table 1 in Appendix C.

Normalization according to a running mean and standard deviation were applied to the observations, actions, and rewards for each EPG inner loop worker independently (Algorithm 1) and for test-time training (Algorithm 2). Adam (18) is used for the EPG inner loop optimization and test-time training with $\beta_1 = 0.9$ and $\beta_2 = 0.999$, while the outer loop ES gradients are modified by Adam with $\beta_1 = 0$ and $\beta_2 = 0.999$ (which means momentum has been turned off) before updating the loss function. Furthermore, L2-regularization over the loss function parameters with coefficient 0.001 is added to outer loop objective. The inner loop step size is fixed to $10^{-3}$, while the outer loop step size is annealed linearly from $10^{-2}$ to $10^{-3}$ over the first 2000 epochs.

### 4.1. Performance

We compare test-time training performance using the EPG loss function, Algorithm 2, against an off-the-shelf policy gradient method, PPO (45). Figures 5, 6, 7, and 11 show learning curves for these two methods on the RandomHopper, RandomWalker, RandomReacher, and Fetch environments respectively at test time. The top plot shows the episodic return w.r.t. the number of environment steps taken so far. The bottom plot shows how much the policy changes at every update by plotting the KL-divergence between the policy distributions before and after every update, w.r.t. the number of updates so far.

In all of these environments, the PPO agent learns by observing reward signals whereas at test time, the EPG agent does not observe rewards (note that at test time, $\alpha$ in Eq. (10) equals 0). Observing rewards is not needed in EPG, since any piece of information the agent encounters forms an input to the EPG loss function. As long as the agent can identify which task to solve within the distribution, it does not matter whether this identification is done through observations or rewards. Keep in mind, however, that the rewards were used in the ES objective function during the EPG evolution phase.

In all experiments, EPG agents learn more quickly and obtain higher returns compared to PPO agents, as expected, since the EPG loss function is able to tailor itself to the environment distribution it is metatrained on. This indicates that our method generates an objective that is more effective at training agents, within these task distributions, than an off-the-shelf on-policy policy gradient method. This is true even though the learned loss does not observe rewards at test time. This demonstrates the potential to use EPG when rewards are only available at training time, for example, if a system were trained in simulation but deployed in the real world where reward signals are hard to measure.

The correlation between the gradients of our learned loss and the PPO objective is around $\rho = 0.5$ (Spearman's rank correlation coefficient) for the environments tested. This indicates that the gradients produced by the learned loss are related to, but different from, those produced by the PPO objective.

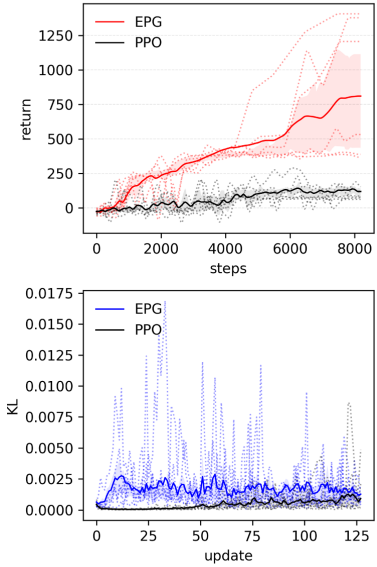
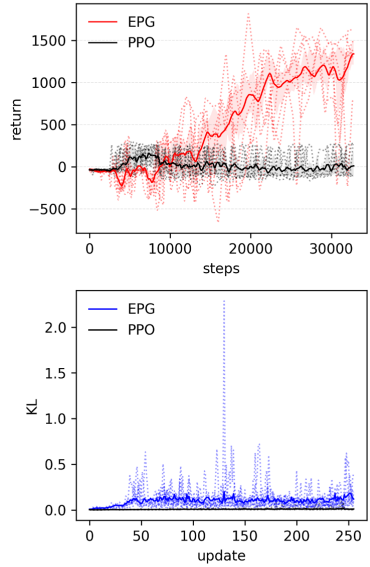
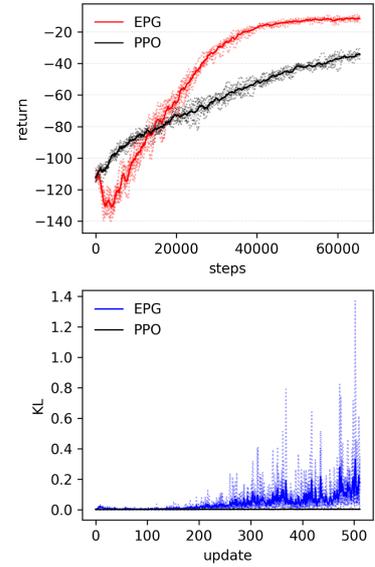

Figure 5: RandomHopper test-time training over 128 (policy updates) ×64 (update frequency) = 8196 timesteps: PPO vs no-reward EPG

Figure 6: RandomWalker test-time training over 256 (policy updates) ×128 (update frequency) = 32768 timesteps: PPO vs no-reward EPG

Figure 7: RandomReacher test-time training over 512 (policy updates) ×128 (update frequency) = 65536 timesteps: PG vs no-reward EPG.

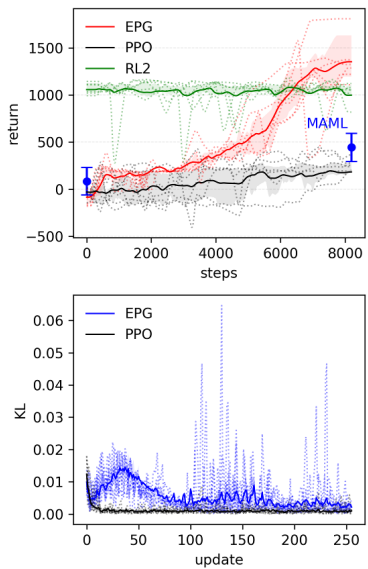
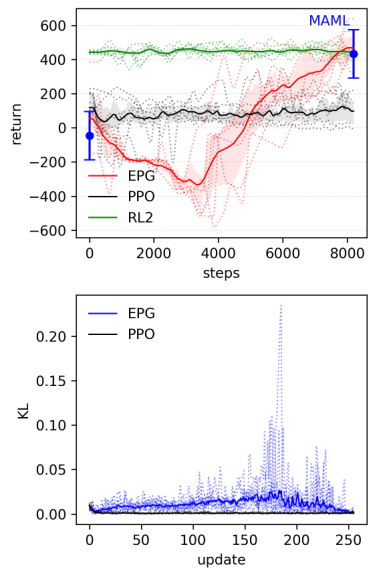
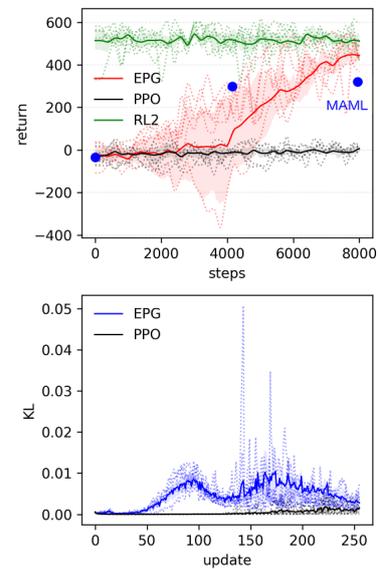

Figure 8: DirectionalHopper environment: each Hopper environment randomly decides whether to reward forward or backward hopping. The agent needs to identify whether to jump forward or backwards: PPO vs EPG. Here we can clearly see exploratory behavior, indicated by the negative spikes in the reward curve, the loss forces the policy to try out backwards behavior. Each subplot column corresponds to a different randomization of the environment.

Figure 9: Comparison with MAML (single gradient step after metalearning a policy initialization) on the DirectionalHalfCheetah environment from Finn et al. [11] (Fig. 5)

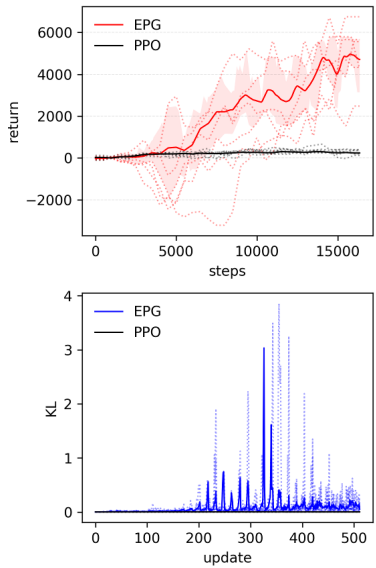
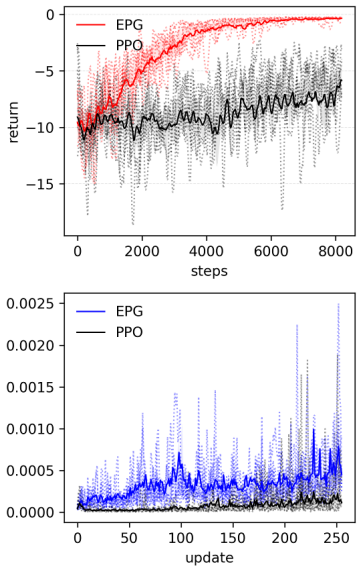
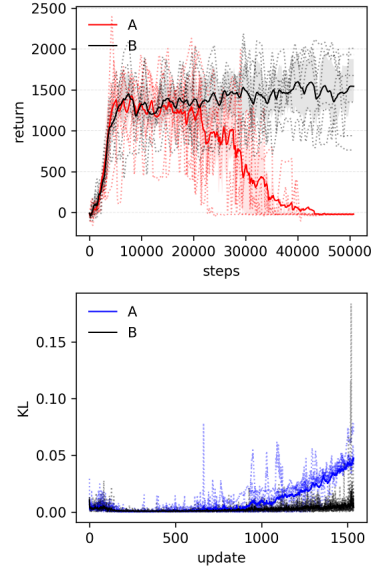

Figure 10: GoalAnt test-time training over 512 (policy updates) ×32 (update frequency) = 16384 timesteps: PPO vs EPG

Figure 11: Fetch reaching environment learning over 256 (policy updates) ×32 (update frequency) = 8192 timesteps: PPO vs no-reward EPG

Figure 12: Transferring EPG (metalearned using 128 policy updates on RandomHopper) to 1536 updates at test time: random policy initialization (A), initialization by sampled previous policies (B)

Figures 8, 9, and 10 show experiments in which a signaling flag is required to identify the environment. Generally, this is done through a reward function or an observation flag, which is why EPG takes the reward as input in case the state space is partially-observed. Similar to the previous experiments, EPG significantly outperforms PPO on the task distribution it is metatrained on. Specifically, in Figure 9, we compare EPG with both MAML (data from (11)) and RL$^2$ (10). This experiment shows that, at least in this experimental setup, starting from a random policy initialization can bring as much benefit as learning a good policy initialization (MAML). In Section 4.2, we will investigate what the effect of evolving the policy initialization together with the loss parameters is. When comparing EPG to RL$^2$ (a method that learns a recurrent policy that does not reset the internal state upon trajectory resets), we see that RL$^2$ solves the DirectionalHalfCheetah task almost instantly through system identification. By learning both the algorithm and the policy initialization simultaneously, it is able to significantly outperform both MAML and EPG. However, this comes at the cost of generalization power, as we will discuss in Section 4.3.

### 4.2. Analysis

In this section, we first analyze whether EPG produces a loss function that encourages exploration and adaptive policy updates during test-time training. Next, we evaluate the effect of evolving the policy initialization.

**Learning exploratory behavior** Without additional exploratory incentives, PG methods lead to suboptimal policies. To understand whether EPG is able to train agents that explore, we test our method and PPO on the DirectionalHopper and GoalAnt environments. In DirectionalHopper, each sampled Hopper environment either rewards the agent for forward or backward hopping. Note that without observing the reward, the agent cannot infer whether the Hopper environment desires forward or backward hopping. Thus we augment the environment reward to the input batches of the loss function in this setting.

Figure 8 shows learning curves of both PPO agents and agents trained with the learned loss in the DirectionalHopper environment. The learning curves give indication that the learned loss is able to train agents that exhibit exploratory behavior. We see that in most instances, PPO agents stagnate in learning, while agents trained with our learned loss manage to explore both forward and backward hopping and eventually hop in the correct direction. Figure 8 (right) demonstrates the qualitative behavior of our agent during learning and Figure 14 visualizes the exploratory behavior. We see that the hopper first explores one hopping direction before learning to hop backwards.

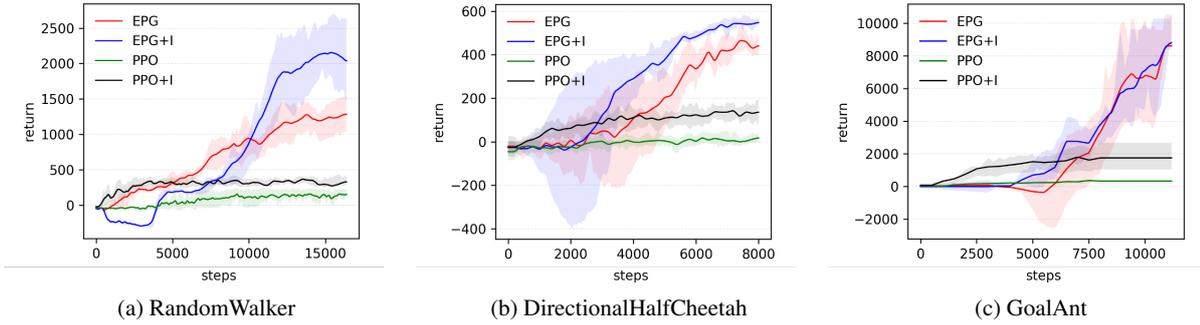

(a) RandomWalker  (b) DirectionalHalfCheetah  (c) GoalAnt

Figure 13: Effect of evolving the policy initialization (+I) on various randomized environments. test-time training curves with evolved policy initialization start at the same return value as those without evolved initialization. This is consistent with MAML trained on a wide task distribution (Figure 5 of (11)).

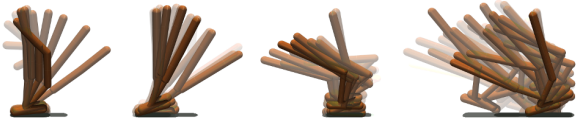

Figure 14: Example of learning to hop backward from a randomly initialized policy in a DirectionalHopper environment. From left to right, trajectories are recorded as learning progresses.

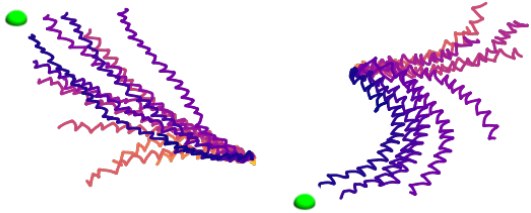

Figure 15: Trajectories sampled from test-time training on two sampled GoalAnt environments: the Ant learns how to explore various directions before going to the correct target. Lighter colors represent initial trajectories, darker colors are later trajectories, according to the agent's learning process.

The GoalAnt environment randomizes the location of the goal. We augment the goal location to the input batches of the loss function. Figure 15 demonstrates the exploratory behavior of a learning ant trained by EPG. We see that the ant first learns to walk and explore various directions, before finally converging on the correct goal location. The ant first explores in various directions, including the opposite direction of the target location. However, it quickly figures out in which quadrant to explore, before it fully learns the correct direction to walk in.

**Learning adaptive policy updates** PG methods such as REINFORCE (60) suffer from unstable learning, such that a large learning step size leads to policy crashing during learning. To encourage smooth policy updates, methods such as TRPO (44) and PPO (45) were proposed to limit the distributional change from each policy update, through a hyperparameter constraining the KL-divergence between the policy distributions before and after each update. We demonstrate that EPG produces learned loss that adaptively scales the gradient updates.

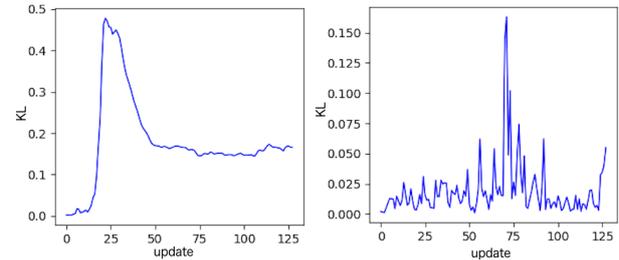

Figure 16: EPG on the RandomHopper environment: the KL-divergence between the policy before and after an update at the first epoch (left) vs the final epoch (right), w.r.t the number of updates so far, for a single inner loop run. These curves are generated with $\alpha = 0$ in Eq. (10).

Figure 16 shows the KL-divergence between policies from one update to the next during the course of training in RandomHopper, using a randomly initialized loss (left) versus a learned loss produced by Algorithm 1 (right). With a learned loss function, the policy updates tend to shift the policy distribution less on each step, but sometimes produce sudden changes, indicated by the spikes. These spikes are highly noticeable in Figure 22 of Appendix B, in which we plot individual test-time training curves for several randomized environments. The loss function has evolved in such a way to adapt its gradient magnitude to the current agent state: for example in the DirectionalHalfCheetah experiments, the agent first ramps up its velocity in one direction (visible by a increasing KL-divergence) until it realizes whether it is going in the right/wrong direction. Then it either further ramps up the velocity through stronger gradients, or emits a turning signal via a strong gradient spike (e.g., visible by the spikes in Figure 22 (a) in column three).

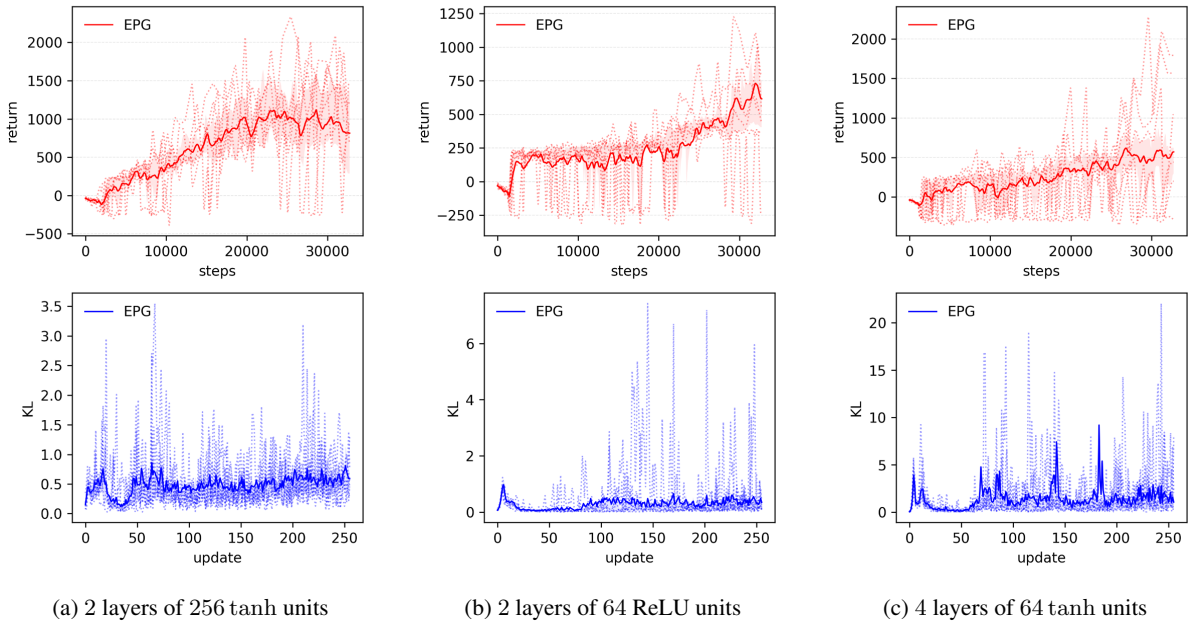

(a) 2 layers of 256 tanh units   (b) 2 layers of 64 ReLU units   (c) 4 layers of 64 tanh units

Figure 17: Transferring EPG (metalearned using 2-layer 64 tanh-unit policies on RandomWalker as in Figure 6) to policies of unseen configurations at test time

In other experiments, such as Figures 8, 9, and 10, we see a similar pattern. Based on the agent's learning history, the gradient magnitudes are scaled accordingly. Often the gradient will be small initially, and it gets increasingly larger the more environment information it has encountered.

**Effect of evolving policy initialization**   Prior works such as MAML (11) metalearn the policy initialization over a task distribution. While our proposed method, EPG, evolves the loss function parameters, we can also augment Algorithm 1 with simultaneously evolving the policy initialization in the ES outer loop. We investigate the benefits of evolving the policy initialization on top of EPG and PPO on our randomized environments. Figure 13 shows the comparison between EPG, EPG with evolved policy initialization (EPG+I), PPO, and PPO with evolved policy initialization (PPO+I). Evolving the policy initialization seems to help the most when the environments require little exploration, such as RandomWalker. However, the initialization plays a far less important role in DirectionalHalfCheetah and especially the GoalAnt environment. Hence the smaller performance difference between EPG and EPG+I.

Another interesting observation is that evolving the policy initialization, together with the EPG loss function (EPG+I), leads to qualitatively different behavior than PPO+I. In PPO+I, the initialization enables fast learning initially, before the return curves saturate. Obtaining a policy initialization that performs well without learning updates was impossible, since there is no single initialization that performs well for all tasks $\mathcal{M}$ sampled from the task distribution $p(\mathcal{M})$. In the EPG case however, we see that the return curves are often lower initially, but higher at the end of learning. By feeding the final return value as the objective function to the ES outer loop, the algorithm is able to avoid myopic return optimization. EPG+I sets the policy up for initial exploratory behavior which, although not beneficial in the short term, improves ultimate agent behavior.

### 4.3. Generalization

Key components of Algorithm 1 include inner-loop training horizon $U$, the agent's policy architecture $\pi_\theta$, and the task distribution $p(\mathcal{M})$. In this section, we investigate the test-time generalization properties of EPG: generalization to longer training horizons, to different policy architectures, and to out-of-distribution tasks.

**Longer training horizons**   We evaluate the effect of transferring to longer agent training periods at test time on the RandomHopper environment by increasing the test-time training steps $U$ in Algorithm 2 beyond the inner-loop training steps $U$ of Algorithm 1. Figure 12 (A) shows that the learning curve declines and eventually crashes past the

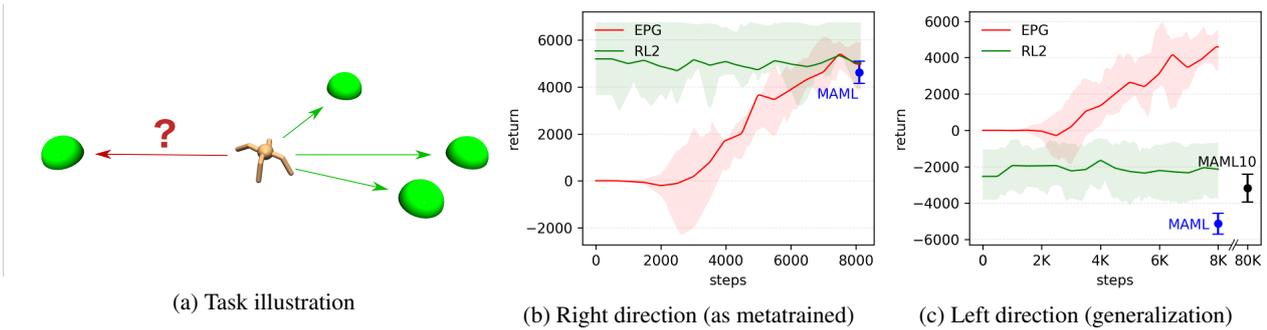

(a) Task illustration  (b) Right direction (as metatrained)  (c) Left direction (generalization)

Figure 18: Generalization in the GoalAnt experiment: the ant has only been metatrained to reach target on the positive x-axis (its right side). Can it generalize to targets on the negative x-axis (its left side)?

train-time horizon, which demonstrates that Algorithm 1 has limited generalization beyond EPG's inner-loop training steps. However, we can overcome this limitation by initializing each inner-loop policy with randomly sampled policies that have been obtained by inner-loop training in past epochs. Figure 12 (B) illustrates continued learning past the train-time horizon, validating that this modification effectively makes the learned loss function robust to longer training length at test time.

**Different policy architectures** We evaluate EPG's transfer to different policy architectures by varying the number of hidden layers, the activation function, and hidden units of the agent's policy at test time (Algorithm 2), while keeping the agent's policy fixed at 2-layer with 64 tanh units during training time (Algorithm 1) on the RandomWalker environment. The test-time training curves on varied policy architectures are shown in Figure 17. Compared to the learning curve Figure 6 with the same train-time and test-time policy architecture, the transfer performance is inferior. However, we still see that EPG produces a learned loss function that generalizes to policies other than it was trained on, achieving non-trivial walking behavior.

**Out-of-distribution task learning** We evaluate generalization to out-of-distribution task learning on the GoalAnt environment. During metatraining, goals are randomly sampled on the positive x-axis (ant walking to the right) and at test time, we sample goals from the negative x-axis (ant walking to the left). Achieving generalization to the left side is not trivial, since it may be easy for a metalearner to overfit to the task metatraining distribution. Figure 18 (a) illustrates this generalization task. We compare the performance of EPG against MAML (11) and $RL^2$ (10). Since PPO is not metatrained, there is no difference between both directions. Therefore, the performance of PPO is the same as shown in Figure 10.

First, we evaluate all metalearning methods' performance when the test-time task is sampled from the training-time task distribution. Figure 18 (b) shows the test-time training curve of both $RL^2$ and EPG when the test-time goals are sampled from the positive x-axis. As expected, $RL^2$ solves this task extremely fast, since it couples both the learning algorithm and the policy. EPG performs very well on this task as well, learning an effective policy from scratch (random initialization) in 8192 steps, with final performance matching that of $RL^2$. MAML achieves approximately the same final performance after taking a single SGD step (based on 8192 sampled steps).

Next, we look at the generalization setting with test-time goals sampled from the negative x-axis. Figure 18 (c) displays the test-time training curves of both methods. $RL^2$ seems to have completely overfit to the task distribution, it has not succeeded in learning a general learning algorithm. Note that, although the $RL^2$ agent still walks in the wrong direction, it does so at a lower speed, indicating that it notices a deviation from the expected reward signal. When looking at MAML, we see that MAML has also overfit to the metatraining distribution, resulting in a walking speed in the wrong direction similar to the non-generalization setting. The plot also depicts the result of performing 10 gradient updates from the MAML init, denoted MAML10 (note that each gradient update uses a batch of 8192 steps). With multiple gradient steps, MAML is able to outperform $RL^2$, consistent with (12), but still learns at a far slower rate than EPG (in terms of number of timesteps of experience required). MAML can achieve this because it uses a standard PG learning algorithm to make progress beyond its init, and therefore enjoys the generalization property of generic PG methods.

In contrast, EPG evolves a loss function that trains agents to quickly reach goals sampled from negative x-axis, never seen during metatraining. This demonstrates rudimentary generalization properties, as may be expected from learning a loss function that is decoupled from the policy. Figure 15 also shows trajectories sampled during the EPG learning process for this exact experimental setup.[2]

## 5. Relation to Existing Literature

The concept of learning an algorithm for learning is quite general, and hence there exists a large body of somewhat disconnected literature on the topic.

To begin with, there are several relevant and recent publications in the metalearning literature (11; 10; 59; 25). In (11), an algorithm named MAML is introduced. MAML treats the metalearning problem as in initialization problem. More specifically, MAML attempts to find a policy initialization from which only a minimal number of policy gradient steps are required to solve new tasks. This is accomplished by performing gradient descent on the original policy parameters with respect to the post policy update rewards. In Section 4.1 of Finn et al. (13), learning the MAML loss via gradient descent is proposed. Their loss has a more restricted formulation than EPG and relies on loss differentiability with respect to the objective function.

In a work concurrent with ours, Yu et al. (62) extended the model from (13) to incorporate a more elaborate learned loss function. The proposed loss involves temporal convolutions over trajectories of experience, similar to the method proposed in this paper. However, unlike our work, (62) primarily considers the problem of behavioral cloning. Typically, this means their method will require demonstrations, in contrast to our method which does not. Further, their outer objective does not require sequential reasoning and must be differentiable and their inner loop is a single SGD step. We have no such restrictions. Our outer objective is long horizon and non-differentiable and consequently our inner loop can run over tens of thousands of timesteps.

Another recent metalearning algorithm is $RL^2$ (10) (and related methods such as (59) and (25)). $RL^2$ is essentially a recurrent policy learning over a task distribution. The policy receives flags from the environment marking the end of episodes. Using these flags and simultaneously ingesting data for several different tasks, it learns how to compute gradient updates through its internal logic. $RL^2$ is limited by its decision to couple the policy and learning algorithm (using recurrency for both), whereas we decouple these components. Due to $RL^2$'s policy-gradient-based optimization procedure, we see that it does not directly optimize final policy performance nor exhibit exploration. Hence, extensions have been proposed such as E-$RL^2$ (53) in which the rewards of episodes sampled early in the learning process are deliberately set to zero to drive exploratory behavior.

Further research on meta reinforcement learning comprises a vast selection. The literature's vastness is further complicated by the fact that the research appears under many different headings. Specifically, there exist relevant literature on: life-long learning, learning to learn, continual learning, and multi-task learning. For example, (41; 40) consider self-modifying learning machines (genetic programs). If we consider a genetic program that itself modifies the learned genetic program, we can subsequently derive a meta-GP approach (See (53), for further discussion on how this method relates to the more recent metalearning literature discussed above). The method described above is sufficiently general that it encompass most modern metalearning approaches. For a further review of other metalearning approaches, see the review articles (48; 57; 58) and citation graph they generate.

There are several other avenues of related work that tackle slightly different problems. For instance, several methods attempt to learn a reward function to drive learning. See (7) (which suggests learning from human feedback) and the field of Inverse Reinforcement Learning (28) (which recovers the reward from demonstrations). Both of these fields relate to our ideas on loss function learning. Similarly, (29; 30) apply population-based evolutionary algorithms to reward function learning in gridworld environments. This algorithm is encompassed by the algorithms we present in this paper. However, it is typically much easier since learning just the reward function is in many cases a trivial task (e.g., in learning to walk, mapping the observation of distance to a reward function). See also (49; 50) and (1) for additional evolutionary perspectives on reward learning. Other reward learning methods include the work of Guo et al. (14), which focuses on learning reward bonuses, and the work of Sorg et al. (51), which focuses on learning reward functions through gradient descent. These bonuses are typically designed to augment but not replace the learned reward and have not been shown to easily generalize across broad task distributions. Reward bonuses are closely linked to the idea of curiosity, in which an agent attempts to learn an internal reward signal to drive future exploration. Schmidhuber (39) was perhaps the first to examine the problem of intrinsic motivation in a metalearning context. The proposed algorithms make use of dynamic programming to explicitly partition experience into checkpoints. Further, there is usually little focus on metalearning the curiosity signal across several different tasks. Finally, the work of (17; 61; 2; 22; 24) studies metalearning over the optimization process in which metalearner makes explicit updates to a parametrized model in supervised settings.

Also worth mentioning is that approaches such as UVFA (38) and HER (3), which learn a universal goal-directed value function, somewhat resemble EPG in the sense that their critic could be interpreted as a sort of loss function that is learned according to a specific set of rules. Furthermore, in DDPG (23), the critic can be interpreted in a similar

---

[2]A demonstration can be viewed at http://blog.openai.com/evolved-policy-gradients/.

way since it also makes use of back-propagation through a learned function into a policy network.

## 6. Discussion

In this paper, we introduced a new metalearning approach capable of learning a differentiable loss function over thousands of sequential environmental actions. Crucially, this learned loss is both adaptive (allowing for quicker learning of new tasks) and instructive (sometimes eliminating the need for environmental rewards at test time), while exhibiting stronger generalization properties than contemporary metalearning methods.

In certain cases, the adaptability of our learned loss is appreciated. For example, consider the DirectionalHopper experiments from Section 4. Here, the rewards at test time are impossible to infer from observations of the environment alone. Therefore, they cannot be completely internalized. However, when we do get to observe a reward signal on these environments, then EPG *does* improve learning speed.

Meanwhile, in most other cases, our loss' instructive nature – which allows it to operate at test time without environmental rewards – is interesting and desirable. This instructive nature can be understood as the loss function's internalization of the reward structures it has previously encountered under the training task distribution. We see this internalization as a step toward learning intrinsic motivation. A good intrinsically motivated agent would successfully infer useful actions in new situations by using heuristics it developed over its entire lifetime. This ability is likely required to achieve truly intelligent agents (39).

Furthermore, through decoupling of the policy and learning algorithm, EPG shows rudimentary generalization properties that go beyond current metalearning methods such as RL$^2$. Improving the generalization ability of EPG, as well other other metalearning algorithms, will be an important component of future work. Right now, we can train an EPG loss to be effective for one small family of tasks at a time, e.g., getting an ant to walk left and right. However, the EPG loss for this family of tasks is unlikely to be at all effective on a wildly different kind of task, like playing Space Invaders. In contrast, standard RL losses do have this level of generality – the same loss function can be used to learn a huge variety of skills. EPG gains on performance by losing on generality. There may be a long road ahead toward metalearning methods that both outperform standard RL methods and have the same level of generality.

Improving computational efficiency is another important direction for future work. EPG demands sequential learning. That is to say, one must first perform outer loop update $i$ before learning about update $i + 1$. This can bottleneck the metalearning cycle and create large computational demands. Indeed, the number of sequential steps for each inner-loop worker in our algorithm is $E \times U$, using notation from Algorithm 1. In practice, this value may be very high, for example, each inner-loop worker takes approximately 196 million steps to evolve the loss function used in the RandomReacher experiments (Figure 7). Finding ways to parallelize parts of this process, or increase sample efficiency, could greatly improve the practical applicability of our algorithm. Improvements in computational efficiency would also allow the investigation of more challenging tasks. Nevertheless, we feel the success on the environments we tested is non-trivial and provides a proof of concept of our method's power.

## Acknowledgments

We thank Igor Mordatch, Ilya Sutskever, John Schulman, and Karthik Narasimhan for helpful comments and conversations. We thank Maruan Al-Shedivat for assisting with the random MuJoCo environments.

## A. Environment Description

We describe the randomized environments used in our experiments in the following:

- *RandomHopper* and *RandomWalker*: randomized gravity, friction, body mass, and link thickness at metatraining time, using a forward-velocity reward. At test-time, the reward is not fed as an input to EPG.

- *RandomReacher*: randomized link lengths, using the negative distance as a reward at metatraining time. At test-time, the reward is not fed as an input to EPG, however, the target location is fed as an input observation.

- *DirectionalHopper* and *DirectionalHalfCheetah*[3]: randomized velocity reward function.

- *GoalAnt*: ant environment with randomized target location and randomized initial rotation of the ant. The velocity to the target is fed in as a reward. The target location is not observed.

- *Fetch*: randomized target location, the reward function is the negative distance to the target. The reward function is not an input to the EPG loss function, but the target location is.

## B. Additional Experiments

**Learning without environment resets** We show that it is straightforward to evolve a loss that is able to perform well on no-reset learning, such that the agent is never reset to a fixed starting location and configuration after each episode. Figure 19 shows the average return w.r.t. the epoch on the GoalAnt environment without reset. The ant continues learning from the location and configuration after each episode finishes and is reset to the starting point only when the target is reached. Qualitative inspection of the learned behavior shows that the agent learns how to reach the target multiple times during its lifetime. In comparison, running PPO in a no-reset environment is highly difficult, since the agent's policy tends to get stuck in a position it cannot escape from (leading to an almost flat zero-return learning curve). In some way, this demonstrates that EPG's learned loss guides the agent to avoid states from which it cannot escape.

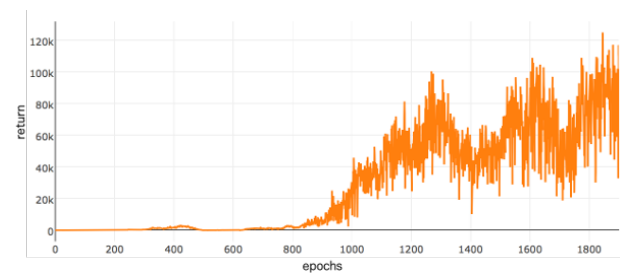

Figure 19: The average return w.r.t. the epoch on the GoalAnt environment with no reset.

**Training performance w.r.t. evolution epoch** Figure 20 shows the metatraining-time performance (calculated based on the noise-perturbed loss functions) w.r.t. the number ES epochs so far, averaged across 256 different inner-loop workers for various random seeds, on several of our environments. This experiment highlights the stability of finding well-performing loss function via evolution. All experiments use 256 workers over 64 noise vectors and 256 updates every 32 steps (8196-step inner loop).

---

[3]Environment sourced from http://github.com/cbfinn/maml_rl.

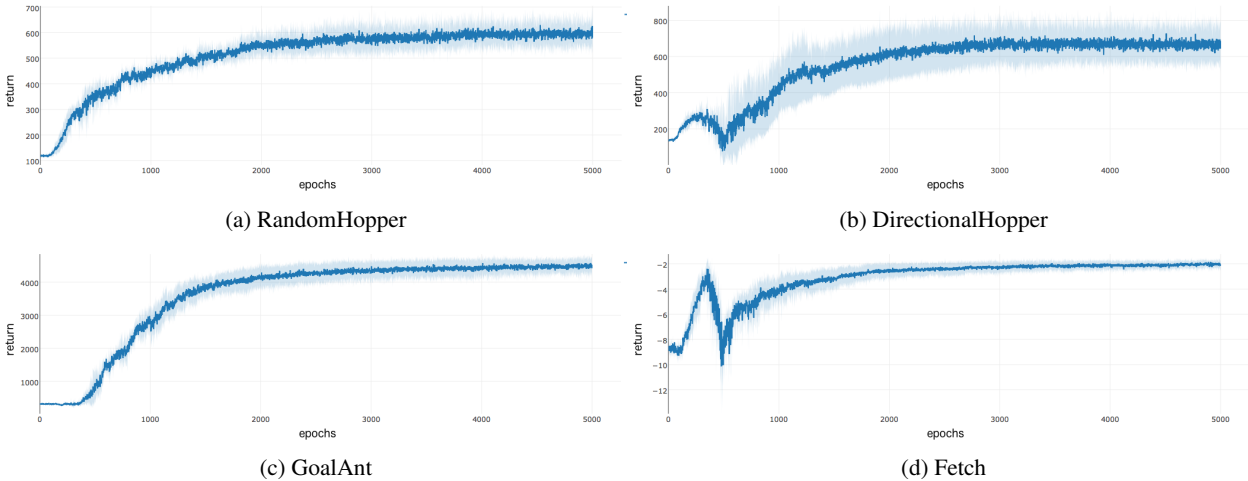

Figure 20: Final returns averaged across 256 inner-loop workers w.r.t. the number outer-loop ES epochs so far in EPG training (Algorithm 1). We run EPG training on each environment across 5 different random seeds and plot the mean and standard deviation as a solid line and a shaded area respectively.

**EPG loss input sensitivity**   In the reward-free case (e.g., RandomHopper, RandomWalker, RandomReacher, and Fetch), the EPG loss function takes four kinds of inputs: observations, actions, termination signals, and policy outputs, and evaluates entire buffer with $N$ transition steps. Which types of input and which time points in the buffer matter the most? In Figure 21, we plot the sensitivity of the learned loss function to each of these kinds of inputs by computing $||\frac{\partial L_{t=25}}{\partial x_t}||_2$ for different kinds of input $x_t$ at different time points $t$ in the input buffer. This analysis demonstrates that the loss is especially sensitive to experience at the current time step where it is being evaluated, but also depends on the entire temporal context in the input buffer. This suggests that the temporal convolutions are indeed making use of the agent's history (and future experience) to score the behavior.

**Individual test-time training curves**   Figures 5, 9, and 10 show the test-time training trajectories of the EPG agent on RandomHopper, DirectionalHalfCheetah, and GoalAnt. A detailed plot of how individual learners behave in each environment is shown in Figure 22. Looking at both the return and KL plots for the DirectionalHalfCheetah and GoalAnt environments, we see that the agent ramps up its velocity, after which it either finds out it is going in the right direction or not. If it is going in the wrong direction initially, it provides a counter signal, turns, and then ramps up its velocity in the appropriate direction, increasing its return. This demonstrates the exploratory behavior that occurs in these environments. In the RandomHopper case, only a slight period of system identification exists, after which the velocity of the hopper is quickly ramped up (visible by the increasing KL divergences).

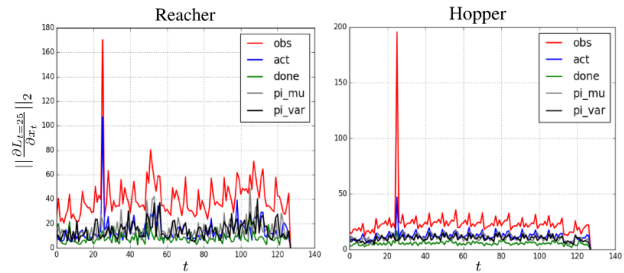

Figure 21: Loss input sensitivity: gradient magnitude of $L_{t=25}$ w.r.t. its inputs at different time steps within the input buffer. Notice not only the strong dependence on current time point ($t = 25$), but also the dependence on the entire buffer window.

## C. Experiment Hyperparameters

The experiment hyperparameters used in Section 4 are listed in Table 1.

| Environment | workers $W$ | noise vectors $V$ | update frequency $M$ | updates | inner loop length |
|---|---|---|---|---|---|
| RandomHopper | 256 | 64 | 64 | 128 | 8196 |
| RandomWalker | 256 | 64 | 128 | 256 | 32768 |
| RandomReacher | 256 | 64 | 128 | 512 | 65536 |
| DirectionalHopper | 256 | 64 | 64 | 128 | 8196 |
| DirectionalHalfCheetah | 256 | 64 | 32 | 256 | 8196 |
| GoalAnt | 256 | 64 | 32 | 512 | 16384 |
| Fetch | 256 | 64 | 32 | 256 | 8192 |

Table 1: EPG hyperparameters for different environments

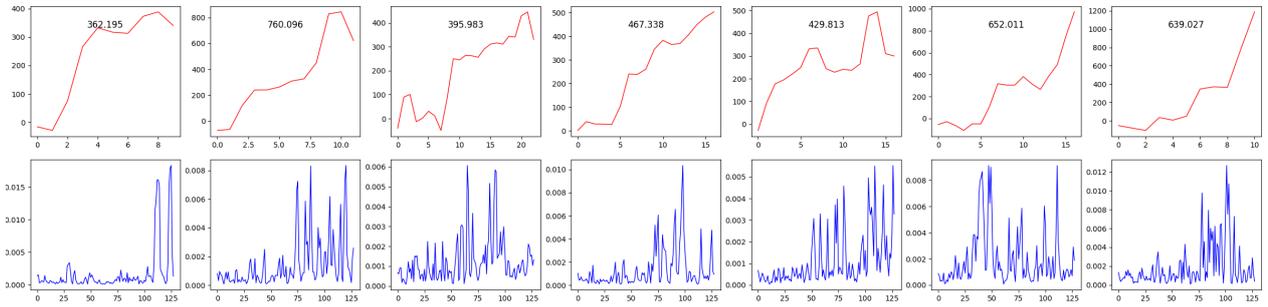

(a) Different runs of the learning agent in Figure 5 (RandomHopper)

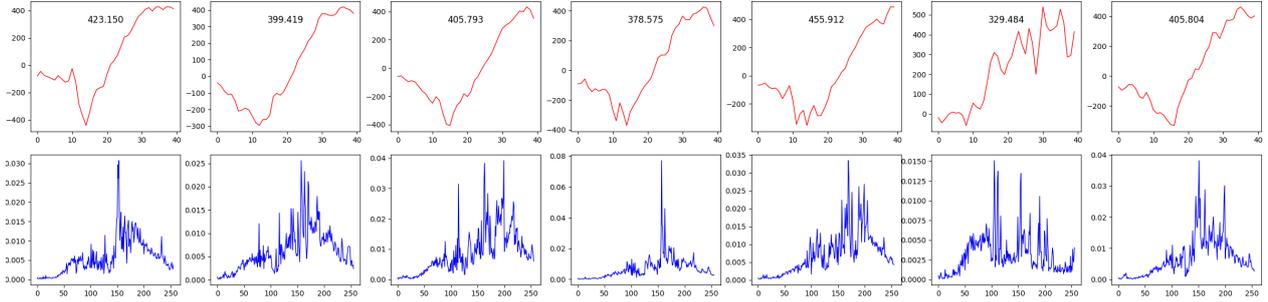

(b) Different runs of the learning agent in Figure 9 (DirectionalHalfCheetah)

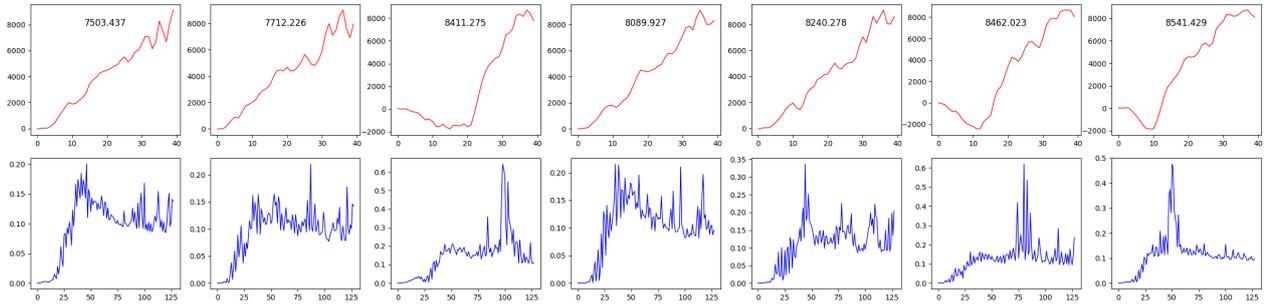

(c) Different runs of the learning agent in Figure 10 (GoalAnt, but limited to forward/backward goals)

Figure 22: More test-time training curves in randomized environments. Each column represents a different sampled environment. The red curves plots the return w.r.t. the number of sampled trajectories during training, while the blue curves represent the KL divergence of the policy updates w.r.t. the number of policy updates. The number shown in each first row plot represents the final return averaged over the final 3 trajectories. The return curve (red) x-axis represent the number of trajectories sampled so far, while the KL-divergence (blue) x-axis represents the number of updates performed so far.